\documentclass[runningheads]{llncs}

\usepackage{hyperref}
\usepackage{graphicx}
\usepackage{amsmath}
\usepackage{amsthm}
\usepackage{booktabs}
\usepackage{algorithm}
\usepackage{algorithmic}
\usepackage{amssymb}
\usepackage{color}
\usepackage{multirow}
\usepackage{latexsym}
\usepackage{bm}
\usepackage{subfigure}
\usepackage[dvipsnames]{xcolor}
\usepackage{colortbl}
\usepackage{balance}
\usepackage{ulem}
\usepackage{setspace}
\usepackage{lipsum}

\newcommand{\chenzhuo}[1]{{\color{black}#1}}

\newcommand{\gyx}[1]{{\color{black}#1}}
\newcommand{\cjy}[1]{{\color{black}#1}}
\newcommand{\jeff}[1]{{\color{black}#1}}

\begin{document}
\title{Zero-shot Visual Question Answering using Knowledge Graph}
\author{
Zhuo Chen\inst{1,2}
\and
Jiaoyan Chen\inst{3}
\and
Yuxia Geng\inst{1,2}
\and
Jeff Z. Pan\inst{4}
\and
Zonggang Yuan\inst{5}
\and
Huajun Chen\inst{1,2}\thanks{Corresponding author.} 
}

\institute{College of Computer Science \& Hangzhou Innovation Center, Zhejiang University
\and
AZFT Joint Lab for Knowledge Engine  \\
\email{\{zhuo.chen,gengyx,huajunsir\}@zju.edu.cn}
\and
Department of Computer Science, University of Oxford\\
\email{jiaoyan.chen@cs.ox.ac.uk}
\and
School of Informatics, The University of Edinburgh
\email{https://knowledge-representation.org/j.z.pan/}
\and
NAIE CTO Office, Huawei Technologies Co., Ltd.\\
\email{yuanzonggang@huawei.com}
}
\authorrunning{Z. Chen et al.}
\maketitle             

\begin{abstract}
\jeff{Incorporating} external knowledge to Visual Question Answering (VQA) has become a vital practical need. 
Existing \cjy{methods mostly adopt pipeline approaches with different components for knowledge matching and extraction, feature learning, etc.
However, such pipeline approaches suffer when some component does not perform well, which leads to error cascading and poor overall performance. 
}
Furthermore, \cjy{the majority of existing approaches ignore the answer bias issue --- many answers may have never appeared  during training (i.e., unseen answers) in real-word application. 
To bridge these gaps, in this paper, we propose a Zero-shot VQA algorithm using knowledge graph and a mask-based learning mechanism for better incorporating external knowledge, and present new answer-based Zero-shot VQA splits for the F-VQA dataset. 
\cjy{Experiments} show that our method can achieve state-of-the-art performance in Zero-shot VQA with unseen answers, meanwhile dramatically augment existing end-to-end models on \cjy{the normal F-VQA task}.}

\end{abstract}

\section{Introduction}
\begin{figure}[htbp]
\centering
\includegraphics[width=1.0\textwidth]{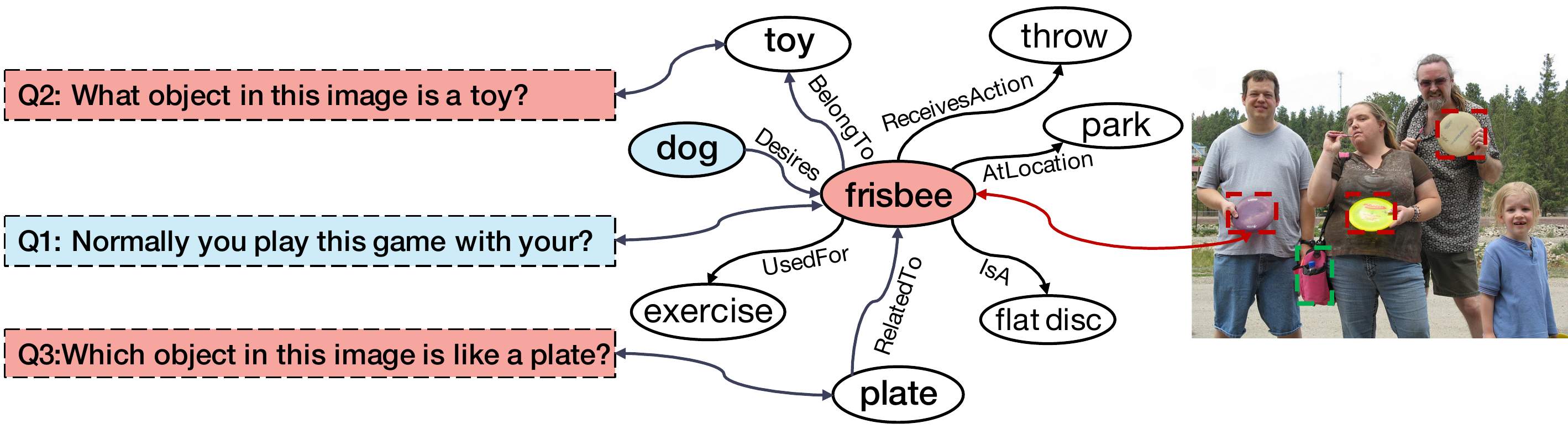}
\caption{\cjy{VQA Examples. Q1: the answer is outside the image and question; Q2 and Q3: the answers are within the images or questions \chenzhuo{but require additional knowledge}.}}
\label{fig:example}
\end{figure}

Visual Question Answering (VQA) is to answer 
\cjy{natural language questions according to given images. 
It plays an important role in many applications such as advertising and personal assistants to the visually impaired. \jeff{It} has been widely investigated with promising results achieved due to the development of image and natural language processing techniques.}
However, most of the current solutions still cannot address the open-world scene understanding where the answer is not directly contained in the image but comes from or relies on external knowledge.
\cjy{Considering} the question ``\cjy{Q1:} Normally you play this game with your?" in Figure \ref{fig:example}, 
some additional knowledge is indispensable since that the answer ``{\it dog}" cannot be found out with the \cjy{content} in the image alone. 

Some VQA methods have been developed to utilize external knowledge for open-world scene understanding. For example, Marino et al. \cite{DBLP:conf/cvpr/MarinoRFM19} extensively \cjy{utilize} unstructured text information from the \cjy{Web as external information but fail to address the noise (irrelevant information) in the text. }
Wang et al. \cite{DBLP:journals/pami/WangWSDH18} first \cjy{extract} visual concepts from images and \cjy{then link} them to an external knowledge graph (KG).  The corresponding questions can then be transformed into a series of \cjy{queries to the KG} (e.g., SPARQL queries) to retrieve \cjy{answers}. 
Zhu et al. \cite{DBLP:conf/ijcai/ZhuYWS0W20} instead construct a multi-modal heterogeneous graph by incorporating the spatial relationships and descriptive semantic relationships between visual concepts, as well as \cjy{supporting} facts retrieved from KGs, 
and then apply a modality-aware graph convolutional network to infer the answer.
However, the performance of all these methods would be dramatically impacted if one module of the pipeline does not perform that well \cjy{(a.k.a. error cascading \cite{DBLP:conf/emnlp/ChenZCXWW20}).}
Although \cjy{some end-to-end models such as \cite{DBLP:conf/nips/KimJZ18,DBLP:conf/cvpr/00010BT0GZ18} have been proposed } to avoid error cascading, they are \cjy{still quite preliminary, especially on utilizing external knowledge, with worse performance than the pipeline methods on many VQA tasks.}

\gyx{Another important \cjy{issue} raised in VQA is the dependence on labeled training data, i.e., the model is trained by a dataset of (question, image, answer) tuples, and generalizes to answer questions about objects and situations that have already been presented in the training set.
However, for new types of questions or answers, and objects newly \cjy{emerge} in images, there is a need for collecting labeled tuples and training the model from the scratch.
Targeting such a limitation, \cjy{Zero-shot VQA (ZS-VQA), which aims to predict with objects, questions or answers that have never appeared in training samples, has been proposed.
Teney et al. \cite{DBLP:journals/corr/TeneyH16a} address questions that include new words; while \cite{DBLP:conf/cvpr/RamakrishnanPSM17,DBLP:journals/ivc/FaraziKB20} address images that contain new objects.}
However, all of these VQA methods still focus on the closed-world scene understanding without considering unseen answers and rarely make full use of KG.
\cjy{In this paper, we utilize KG to study VQA with open-world scene understanding, which requires external knowledge to answer the question, and ZS-VQA, especially the sub-task that addresses new answers}.}

\cjy{In this paper, we present a ZS-VQA algorithm using KG and a mask-based learning mechanism, and at the same time propose a new Zero-shot Fact VQA (ZS-F-VQA) dataset which is to evaluate ZS-VQA for unseen answers. }
\jeff{Firstly, we} learn three different feature mapping spaces separately, which are semantic space about \jeff{relations}, object space about support \jeff{entities}, and knowledge space about \jeff{answers}. Each of them is used to align the joint embedding of image-question pair (I-Q pair) with corresponding target.
Via the combination between all those chosen supporting entities and relations, masks are decided according to a mapping table which contains all triplets in \jeff{a} fact KG, \jeff{which guides} 
the alignment process of unseen answer prediction. 
Specially,
\jeff{the marks can be used as hard masks or soft masks, depending on the VQA tasks.} 
\jeff{Hard marks are used in ZS-VQA tasks; e.g., with the  ZS-F-VQA  dataset,} 
our method achieves state-of-the-art performance and far superior ($30 \sim 40\%$) to other methods.
\jeff{On the other hand, soft marks are used in standard VQA tasks; e.g., with the F-VQA dataset, } 
our method achieves a stable improvement ($6 \sim 9\%$) on baseline end-to-end method and well \cjy{alleviates} the error cascading problem of pipeline models.
\jeff{To sum up, } the main contributions are summarized below:
\begin{itemize}
    \item We propose a \cjy{robust ZS-VQA algorithm using KGs\footnote{\jeff{Our} code and data are available at
\url{https://github.com/China-UK-ZSL/ZS-F-VQA}}}, which adjusts answer prediction score via masking based on the alignments between supporting \jeff{entities/relations} and fusion I-Q pairs in two feature spaces. 
    \item \cjy{We define a new ZS-VQA problem which requires external knowledge and considers unseen answers.
    Accordingly, we develop a ZS-F-VQA dataset for evaluation.}
    \item \cjy{Our KG-based ZS-VQA algorithm is quite flexible. It can successfully address both normal VQA tasks that rely on external knowledge and ZS-VQA tasks, and can be directly used to augment existing end-to-end models.}
\end{itemize}
\section{Related Work} 
\subsection{Visual Question Answering}\label{sec:vqa}
\noindent\textbf{Traditional VQA Methods.} 
\cjy{Since proposed in 2015 by \cite{DBLP:conf/iccv/AntolALMBZP15}, a few VQA methods, which apply multi-modal feature fusion between question and image for final answer decision, have been proposed.}
Various attention mechanisms \cite{DBLP:conf/cvpr/YangHGDS16,DBLP:conf/cvpr/00010BT0GZ18} are adopted to refine specific regions of the image for corresponding question meanwhile \cjy{to} make the prediction process interpretable. 
Graph-based \cjy{approaches such as} \cite{DBLP:conf/icml/ChenG0LC020} combine multi-modal information and enhance the interaction among significant entities in texts and images. 

\noindent\textbf{Knowledge-based VQA.}  
\cjy{Utilizing symbolic knowledge is a straight forward solution to augment VQA.
To study incorporating external knowledge with VQA, datasets such as F-VQA \cite{DBLP:journals/pami/WangWSDH18}, OK-VQA \cite{DBLP:conf/cvpr/MarinoRFM19} and KVQA \cite{DBLP:conf/aaai/ShahMYT19} have been developed}. 
Each question in F-VQA refers to a specific fact triple in relevant KG like ConceptNet.
While OK-VQA is manually marked without a guided KG as reference which leads to its difficulty. 
KVQA targets at world knowledge where questions \jeff{are}   about the relationship between characteristics.

To incorporate such external knowledge, \cite{DBLP:conf/ijcai/WangWSDH17,DBLP:journals/pami/WangWSDH18} generate SPARQL \cjy{queries} for querying the constructed sub-KG according to I-Q pairs. \cite{DBLP:conf/cvpr/WuWSDH16,DBLP:conf/nips/NarasimhanLS18,DBLP:conf/eccv/NarasimhanS18,DBLP:conf/ijcai/ZhuYWS0W20} extract entities \cjy{from} image and question to get related concepts from KG for answer prediction. 
Marino et al.\cite{DBLP:conf/cvpr/MarinoRFM19} take unstructured knowledge on \cjy{the} Web to enhance the semantic representation of I-Q joint feature. 
All of the above 
\cjy{methods utilize} pipeline \cjy{approaches} to narrow the answer scope, but \cjy{they are often ad-hoc, which limits their deployment and generalization to new datasets.}
Most importantly, the errors will be magnified during running since each module usually has no ability to correct previous modules' errors.
End-to-end model like \cite{DBLP:conf/nips/KimJZ18,DBLP:conf/cvpr/00010BT0GZ18} \cjy{are more general and} can avoid error cascading, but they are \cjy{still preliminary, especially in addressing VQA problems} which require external knowledge.

Different from these approaches, our proposed framework leverages \cjy{the advantages of both end-to-end and pipeline approaches}. We improve the model transferability meanwhile \cjy{effectively avoid} the error cascading (see our case study as illustrated in Figure~ \ref{fig:example of general}), \cjy{making it quite general to different tasks and very robust with promising performance achieved.}
\subsection{Zero-shot VQA}\label{tab:Zero-shot VQA}
 Machine learning often follows a closed world assumption \cjy{where classes to predict all have training samples.}
 However, the real-world is not completely closed and it is impractical to always annotate sufficient samples to re-train the model for new classes. Targeting such a limitation, zero-shot learning (ZSL) is proposed to handle these novel classes without seeing their training samples (i.e., unseen classes) \cite{DBLP:conf/www/GengC0PYYJC21,chen2021knowledge}.
Teney et al. \cite{DBLP:journals/corr/TeneyH16a} first propose Zero-shot VQA (ZS-VQA) and introduce novel concepts on language semantic side, where a test sample is regarded as unseen if there is at least one novel word in its question or answer. 
Ramakrishnan et al. \cite{DBLP:conf/cvpr/RamakrishnanPSM17} incorporate prior knowledge into model through pre-training with unstructured external data (from both visual and semantic level). 
Farazi et al. \cite{DBLP:journals/ivc/FaraziKB20} 
 reformulate ZS-VQA as a transfer learning task that applies closely seen instances (I-Q pairs) for reasoning about unseen concepts. 
A major limitation of these approaches is that they seldom consider the imbalance and low resources problem regarding the answer itself.
Open-answer VQA requires  models to select answer with the highest activation from fixed possible $K$ answer categories, but the model cannot tackle unseen answers because answers are isolated with no specific meaning. Besides, VQA is defined as a classification problem without utilizing enough semantic information of the answer.
Agrawal et al. \cite{DBLP:journals/corr/AgrawalKBP17} propose a new setting for VQA where the test question-answer pairs are compositionally novel compared to training question-answer pairs.
Some methods \cite{DBLP:conf/cvpr/HuCS18,DBLP:journals/corr/abs-2005-01239} try to align answer with I-Q joint embedding through feature representation for realizing unseen answer prediction or simply for concatenating their representation as the input of a fully connected layer for score prediction \cite{DBLP:journals/corr/TeneyH16a}. 
However, all of them are powerless to answer those I-Q pairs that require external knowledge, and the relevance among answers are still not strong enough with insufficient external information. 
\cjy{The ZS-VQA method proposed in this paper incorporates richer and more relevant knowledge by using KGs, through which the existing common sense is well utilized and more accurate answers are often given.}
\section{Preliminaries}
\noindent\textbf{Visual Question Answering (VQA) and Zero-shot VQA.} 
A VQA task is  to \jeff{provide} a correct answer $a$ given an image $i$ paired with a question $q$. 
Following the open-answer VQA setting defined in \cite{DBLP:conf/cvpr/HuCS18},   let $a$ be a member of the answer pool $\mathcal{A}=\{a_1, ...,a_n\}$, the candidates of which are the top K (e.g. $500$) most frequent answers of the whole dataset. 
A dataset is represented by a set of distinctive triplets $\mathcal{D} = \{(i, q, a) | i \in \mathcal{I}, q \in \mathcal{Q}, a \in \mathcal{A} \}$ where $\mathcal{I}$ and $\mathcal{Q}$ are respectively image and question sets.
\jeff{A} testing dataset is denoted as $\mathcal{D}_{te}$ with \cjy{each triplet $(i, q, a)$ not belonging to training dataset $\mathcal{D}_{tr}$.}
We denote $\mathcal{D}_{tr}^{zsl} = \{(i, q, a) | i \in \mathcal{I}, q \in \mathcal{Q}, a \in \mathcal{A}_s \}$ and $\mathcal{D}_{te}^{zsl} = \{(i, q, a) | i \in \mathcal{I}, q \in \mathcal{Q}, a \in \mathcal{A}_u \}$,
\cjy{where $\mathcal{A}_s$ and $\mathcal{A}_u$ respectively denote the seen answer set and the unseen answer set with $\mathcal{A}_u \cap \mathcal{A}_s = \emptyset$.}
\cjy{ZS-VQA is much harder than normal VQA, since information in the image and question is insufficient for answers that have never appeared in the training samples.}
Specifically, 
\cjy{we study two settings at testing stage of ZS-VQA : one is the standard ZSL, where the candidates answers of a testing sample $(i, q, a)$ are $\mathcal{A}_u$, 
while the other is the generalized ZSL (GZSL) with $\mathcal{A}_u \cup \mathcal{A}_s$ as candidates answers during testing. }
It should be noted that regular VQA only predicts with seen answers, while VQA under the GZSL setting predicts with both seen and unseen answers.

\noindent\textbf{Knowledge Graph (KG). }
KGs have been widely used in knowledge representation and knowledge management \cite{PVGW2017,PCEH+2017} .
The KG we used is a subset of three KGs (DBpedia, ConceptNet, WebChild) \jeff{selected} by Wang et al. \cite{DBLP:journals/pami/WangWSDH18} (in the form of RDF\footnote{https://www.w3.org/TR/2014/REC-rdf11-mt-20140225/} triple). It is used to establish the prior knowledge connection,
which includes a set of answer nodes and concept (tool) nodes to enrich the relationships among answers. 
Besides, different \cjy{relations (edges)} are applied to represent the \cjy{fact graph (triples).}

Taking  Figure~\ref{fig:example} as an example, all $(i,q)$ pairs could be divided into two categories according to their answer sources :
1) Those answers which are outside the images and questions. Such as the answer ``dog" to question ``Q1: Normally you play this game with your?", the data source of the answer here is the external KG which contains the triple $<$frisbee, BelongTo, toy$>$ for QA support.
2) Those answers that can be found in images or questions. In this situation, there are often more than one object in image/question for screening through some implicit common sense relations (e.g., ``Q2: Which object in this image is like a plate?" targets at finding the correct object related to plate ). Then, one fact triple (e.g. $<$plate, RelatedTo, frisbee $>$) could play the role of answer guidance.

\section{Methodology}
\subsection{Main Idea}
Our main idea is motivated by two deficiencies in current knowledge-based VQA approaches. Firstly, in those methods it is common to build intermediate models and involve KG queries in a pipeline way, which leads to error cascading and poor generalization.
Secondly, most of them define VQA as a classification problem which does not utilize enough knowledge of the answers, and fails to predict unseen answers or to transfer across datasets whose candidate answers have little or no overlap.
For example, as shown in Figure \ref{fig:example}, if concept ``frisbee" has not appeared in training set, traditional VQA will fail to recognize it in testing phase for answer out-of-vocabulary (OOV) problem. While other method \cite{DBLP:conf/cvpr/HuCS18} which takes answer semantics into account has lost the relation information: ``Desires" came from entity ``dog", or ``RelatedTo" came from entity ``plate".
\begin{figure}[htbp]
\centering
\includegraphics[width=1.0\textwidth]{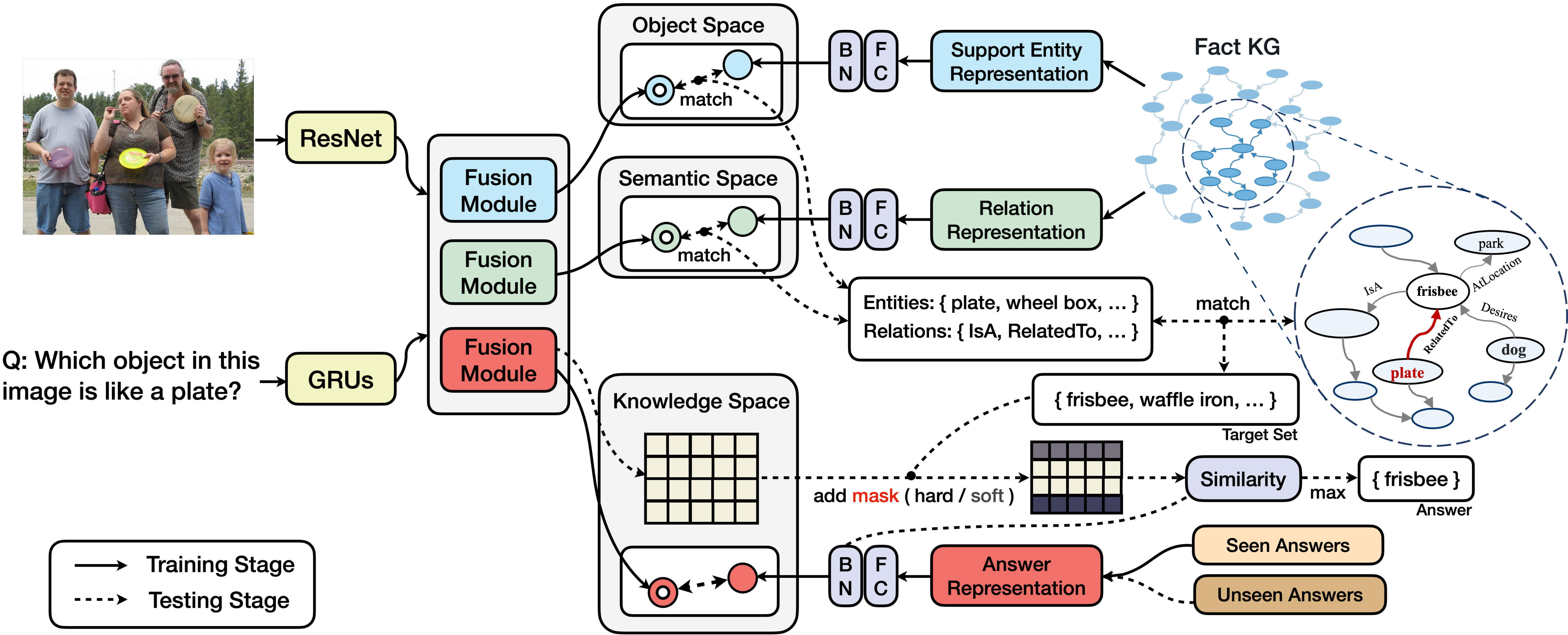} 
\caption{ An overview of our framework.
}
\label{fig:Model_architecture}
\end{figure}

By utilizing semantics embedding feature as answer representation, we convert VQA from a classification task into a mapping task. After parameter learning, the distribution of the joint embedding between question and image can partly get close to answer's one  with shadow knowledge included in. We call it the knowledge space about answers.
Besides, we independently define two other feature spaces: semantic space about relations and object space about support entities.
Semantic space aims to project $(i,q)$ joint feature into a relation according to the semantic information in triplets, while object space targets at establishing relevant connection between  $(i,q)$ and a support entity (a.k.a. entity on KG ).
They play the role for answer guidance when combined together (see Section~\ref{sec:feature Spaces} for detail).
In order to overcome those limitations proposed in Section~\ref{sec:vqa}, we provide a soft/hard mask method in this situation to effectively enhance alignment process meanwhile alleviating error cascading.

\subsection{Establishment of Multiple Feature Spaces} \label{sec:feature Spaces}
Following \cite{DBLP:conf/cvpr/HuCS18}, we establish connection between an answer and its corresponding $(i, q)$ pair via projecting them into a common feature space and get close to each other.
Firstly, a fusion feature extractor $F_{\boldsymbol{\theta}}\left(i, q\right)$ between $q$ and $i$ is leveraged to combine multimodal information.
Meanwhile, we define  $G_{\phi} (a)$
as the representation of answer $a$.
We follow the probabilistic model of compatibility (PMC) drawn from \cite{DBLP:conf/cvpr/HuCS18} and add loss temperature $\tau$ for better optimization:
\begin{equation}
	P\left(a \mid i_{n}, q_{n}\right)=\frac{\exp \left(F_{\boldsymbol{\theta}}\left(i_{n}, q_{n}\right)^{\top} G_{\phi}(a)/\tau\right)}{\sum_{a^{\prime} \in \mathcal{A}} \exp \left(F_{\boldsymbol{\theta}}\left(i_{n}, q_{n}\right)^{\top} G_{\phi}\left(a^{\prime}\right)/\tau\right)}
\end{equation}
where $\mathcal{A}$ denotes $\mathcal{A}_{u}$ when the setting is standard ZSL else remain the same, and $a$ is the correct answer of $(i_n,q_n)$.
For learning the parameters to maximize the likelihood in overall PMC model, we employ following loss function:
\begin{equation}
	\ell_{a}=-\sum_{n}^{N} \sum_{b \in \mathcal{A}} \alpha(a, b) \log P\left(b \mid i_{n}, q_{n}\right)
\end{equation}
where weighting function $\alpha(a, b)$ measures how much the predicted answer $b$ can contribute to the objective function. A nature design is
$\alpha(a, b)=\mathbb{I}[a=b]$, 
where $\mathbb{I}[.]$ denotes binary indicator function, taking value of $1$ if the condition is true else $0$ for false. 
During testing, 
given the learned $F_{\boldsymbol{\theta}}\left(i, q\right)$ and $G_{\phi}(a)$, 
we can apply following decision rule to predict the answer $\hat{a}$ to $(i, q)$ pair:
\begin{equation}
\hat{a}=\arg \max _{a \in \mathcal{A}} F_{\boldsymbol{\theta}}(i, q)^{\top} G_{\boldsymbol{\phi}}(a)	
\end{equation}

Like the results shown in Section~\ref{sec:Overall Results}, the above feature projection process could learn shallow knowledge in VQA which requires external knowledge. 
However, it performs not well since network is not sufficient to model abundant prior knowledge with small amount of training data (see data statistics in Table~\ref{tab:dateset split}).

Matching the elements in images or questions to KG entities in an explicit \cite{DBLP:journals/pami/WangWSDH18} or implicit \cite{DBLP:journals/ivc/FaraziKB20} way can augment the model with knowledge to well address the open-world scene understanding problem (see links in Figure \ref{fig:example} toy example). In our method, the alignment between image/question and KG is implicitly done by multiple feature spaces rather than simply object detection. We leverage another two feature spaces for answer revising:

1) \textbf{Semantic space} focuses on the language information within $(i, q)$, which works as a guidance toward the projection of triplet relations $r$ in KG. In particular, the signal of $q$ is more crucial than $i$ in this part.

2) Compared with traditional image classification which identifies the correct category of a given image, the \textbf{object space} is more likely a feature space about support entity classifier which simultaneously observes images and texts for salient feature. 
Specifically, the alignment between support entity $e$ embedding and  $(i, q)$ joint embedding avoids the direct learning of complex knowledge, meanwhile acts on the subsequent answer mask process together with the prediction relations $r$ obtained in semantic space. 

Similarly, we define their embedding function as $G_{\phi\star}(r)$, $G_{\phi\diamond}(e)$ and the corresponding  $(i, q)$ joint embedding function as $F_{\boldsymbol{\theta\star}}\left(i, q\right)$, $F_{\boldsymbol{\theta\diamond}}\left(i, q\right)$ for distinction.
Other formulas and probability calculation methods remain the same as answer such as loss function $\ell_{r}$ and $\ell_{e}$, which are model's overall optimization goal together with $\ell_{a}$. The parameters in these three pairs of models are independent except for the frozen input embedding layers.

Pre-trained word vector contains the latent semantics in real-world natural language. In order to get the initialized representation of the answer, relation and support entity, we employ GloVe embedding \cite{DBLP:conf/emnlp/PenningtonSM14} meanwhile  compare other answer representation like KG embedding \cite{DBLP:conf/nips/BordesUGWY13} or ConceptNet embedding \cite{DBLP:conf/aaai/MalaviyaBBC20} (see Section~\ref{sec:ablation study} for detail).

Besides,  
\jeff{different surface forms (e.g., mice \& mouse) should be considered for the same meaning}. 
\cite{DBLP:conf/cvpr/HuCS18} takes advantage of the weighting function $\alpha(a, b)$ with WUPS score, which is reliant on semantic similiarity scores between $a$ and $b$. We find that it works well with singular and plural disambiguation (e.g. WUPS(~{\it dog}, {\it dogs}~)~$\approx0.929$), but fails in many cases of tense disambiguation (e.g., WUPS(~{\it cook}, {\it cooking}~)~$\approx0.125$, WUPS(~{\it play}, {\it played}~)~$\approx0.182$). 
So we apply NLTK tools (e.g., WordNetLemmatizer) to achieve more accurate word split and Minimum Edit Distance (MED) for concept disambiguation.
\subsection{Answer Mask via Knowledge}
\cjy{Masking is widely used in language model pre-training for improving machine's understanding of the text. Two examples are masking part of the words in the training corpus
 (e.g. BERT \cite{DBLP:conf/naacl/DevlinCLT19}) and} masking common sense concepts (e.g. AMS \cite{DBLP:journals/corr/abs-1908-06725}). 
But they rarely consider the direct constraint of knowledge in prediction results, \cjy{ignoring that }human beings know how to make reasonable decision under the guidance of existing prior knowledge.
Different from all these methods, we propose an answer masking strategy for VQA.

With the learned $F_{\boldsymbol{\theta\star}}$ and $F_{\boldsymbol{\theta\diamond}}$, we get the disjoint fusion embedding in two independent feature spaces, which are respectively taken as the basis for subsequent entity and relation matching:  
For a given $(i,q)$ pair, vector similarity $Sim$ is calculated  via $F_{\boldsymbol{\theta\star}}(i, q)^{\top} G_{\boldsymbol{\phi\star}}(r_n)$ for relation, and $F_{\boldsymbol{\theta\diamond}}(i, q)^{\top} G_{\boldsymbol{\phi\diamond}}(e_n)$ for support entity. Those $e$ and $r$, which correspond to the top-$k$ $Sim$ value, separately constitute the candidate set $\mathcal{C}_{ent}$ and $\mathcal{C}_{rel}$ 
where $k$ is distinguished with $k_r$ and $k_e$.
Then target set $\mathcal{C}_{tar}$ is collected  as follows:
\begin{equation}
	\mathcal{C}_{tar} = \{t \mid (\exists (t,r,e)\lor\exists (e,r,t))\wedge r\in \mathcal{C}_{rel} \wedge e \in \mathcal{C}_{ent} \}
\end{equation}

$\mathcal{C}_{tar}$ contributes to the masking strategy on all answers $a_n \in \mathcal{A}$ via:
\begin{equation}
sim(\left(i,q),a_n\right)=\left\{\begin{array}{ll}
(F_{\boldsymbol{\theta}}(i, q)^{\top} G_{\boldsymbol{\phi}}(a_n))/\tau + s & \text { if } a_n\in \mathcal{C}_{tar}\\
(F_{\boldsymbol{\theta}}(i, q)^{\top} G_{\boldsymbol{\phi}}(a_n))/\tau  & \text { otherwise }
\end{array}\right.
\end{equation}
where $s$ represents the score for masking which is the mainly distinction between hard mask and soft mask (see Section~\ref{sec:Impact of mask score} for detail). Soft score greatly reduces the error cascading caused by the pipeline method through the whole model, which will be discussed in Section~\ref{sec:Interpretability}. Meanwhile, the significance of hard mask comes from its superior performance in ZSL setting as shown in Section~\ref{sec:Overall Results}.
Finally, the predicted answer $\hat{a}$ to the $(i, q)$ pair is identified as:
\begin{equation}
\hat{a}=\arg \max _{a \in \mathcal{A}} sim(\left(i,q),a\right)	
\end{equation}

It should be noted that candidate targets cannot just be regarded as the candidate answers due to the existence of soft mask, which revises the answer probability rather than simply limits answer's range. 
Moreover, as mentioned in Section~\ref{sec:Impact of fr} and \ref{sec:Impact of mask score}, $k$ and $s$ mentioned above are hyper parameters which can cause various influence toward the result.
\section{Experiments}
We validate our approach
for both normal VQA and ZS-VQA with \cjy{ZSL/GZSL settings. 
In addition to the overall results, we conduct ablation studies for analyzing the impact of: 1) different factors in answer embedding; 2) the mask score; and 3) different \cjy{hyper parameters} (e.g. $k_{e}$, $k_{r}$).}
Finally, we evaluate its advantage on data transferability and mitigating error cascading. 
\subsection{Datasets and Metrics}
\textbf{F-VQA.}
 As a standard publicly available VQA benchmark which requires external knowledge, F-VQA \cite{DBLP:journals/pami/WangWSDH18} consists of $2,190$ images, $5,286$ questions and a KG of $193,449$ facts. Each $(i,q,a)$ in this dataset is supported by a corresponding common sense fact triple extracted from public structured databases (e.g., ConceptNet, DBPedia, and WebChild). The KG has 101K entities and 1833 relations in total, 833 entities are used as answer nodes. 
In order to achieve parallel comparison, we maintain the coincide experiment setting with \cite{DBLP:journals/pami/WangWSDH18,DBLP:conf/eccv/NarasimhanS18} to use standard dataset setting which contains $5$ splits (by images), and prescribe candidate answers to the top-$500$ (\%$94.30$ to entire as our check) for experiments. 
The over all data statistics after disambiguation are shown in Table~\ref{tab:dateset split}.  

\noindent\textbf{ZS-F-VQA.}
\jeff{The} ZS-F-VQA dataset  is a new split of the F-VQA dataset for zero-shot problem.
Firstly we obtain the original train/test split of F-VQA dataset and combine them together to filter out the triples whose answers appear in top-500 according to its occurrence frequency.
Next, we randomly divide this set of answers into new training split (a.k.a. seen) $\mathcal{A}_s$ and testing split (a.k.a. unseen) $\mathcal{A}_u$ at the ratio of 1:1. 
With reference to F-VQA standard dataset, the division process is repeated 5 times. 
For each $(i,q,a)$ triplet in original F-VQA dataset, it is divided into training set if $a \in \mathcal{A}_s$. Else it is divided into testing set.
The data statistics are shown in Table~\ref{tab:dateset split}, where \#class represents the number of data after deduplicated and \#instance represents the number of samples. We denote ``Overlap" as the intersection size of element sets within training and testing triples. 
Note that the overlap of answer instance between training and testing set in F-VQA are $2565$ compared to $0$ in ZS-F-VQA. 
\setlength{\tabcolsep}{2pt}
\begin{table}[htbp]   
\caption{The detailed data statistics. Average number of $(i,q,a)$ in each train/test split in F-VQA is $2757$/$2735$ compared to $2732$/$2760$ of ZS-F-VQA.} 
\label{tab:dateset split}
\scriptsize 
\centering
\begin{tabular}{c|cccc}    
 \toprule
 &   {\bf  Images} & {\bf Question} & {\bf  Answer} & {\bf  Support Entity}\\
\midrule 
$\#$class& Train/Test/Overlap &   Train/Test/Overlap &  Train/Test/Overlap &  Train/Test/Overlap\\
\midrule 
F-VQA & $1059$~/~$1064$~/~\textbf{0} & $2431$~/~$2409$~/~$573$ & $387$~/~$401$~/~$288$ & $1695$~/~$1668.8$~/~$312$ \\ 
ZS-F-VQA & $1297$~/~$1312$~/~$486$ & $2384$~/~$2380$~/~$264$ & $250$~/~$250$~/~\textbf{0} & $1578$~/~$1477$~/~$86$ \\ 
\midrule 
$\#$instance& Overlap &  Overlap &  Overlap &  Overlap\\
\midrule 
F-VQA & \textbf{0} & $1372$ & $2565$ & $312$\\ 
ZS-F-VQA & $990$ & $814$ & \textbf{0} & 218\\ 
 \bottomrule
\end{tabular}  
\end{table}

\noindent\textbf{Evaluation Metrics.} 
We measure the performance by accuracy and choose $Hit@1$, $Hit@3$, $Hit@10$ here together with MRR/MR to judge the comprehensive performance of model. $Hit@X$ indicates that the correct answer ranks in the top-k predicted answer sorted by probability. Mean Reciprocal Rank (MRR) measure the average reciprocal values of correct predicted answers compared to Mean Rank (MR). All the results we report are averaged across all splits.
\subsection{Implementation Details}\label{sec:Experimental Setup}
\textbf{Fusion Model.} We employ several models to parameterize the fusion function $F_{\boldsymbol{\theta}}$. 
We follow \cite{DBLP:conf/cvpr/HuCS18} to employ the Multi-layer Perceptron (MLP) and Stacked Attention Network (SAN) \cite{DBLP:conf/cvpr/YangHGDS16} as the representation of grid based visual fusion model. Meanwhile, we choose Up-Down (Bottom-Up and Top-Down Attention) \cite{DBLP:conf/cvpr/00010BT0GZ18} and Bilinear Attention Networks (BAN) \cite{DBLP:conf/nips/KimJZ18} to measure the impact of bottom-up issue on external knowledge VQA problem. Moreover, we directly compare with \cite{DBLP:journals/pami/WangWSDH18} in some baselines like Qqmaping \cite{DBLP:journals/pami/WangWSDH18}  Hie \cite{DBLP:conf/nips/LuYBP16} under identical setting. 
Among all these methods, SAN is chosen as the base feature extractor $F_{\boldsymbol{\theta}}$ of our framework for its better performance(see Figure~\ref{tab:Overall Results of General}).

\noindent\textbf{Visual Features.} 
To get $i_{n}$, we extract visual features from the layer $4$ output of ResNet-152 ($14\times14\times2048$ tensor) pre-trained on ImageNet.
Meanwhile applying ResNet-101-based Faster R-CNN pre-trained on COCO dataset to get bottom-up image region features. The object number per image is fixed into $36$ with $1024$ output dimensional feature. 

\noindent\textbf{Text Features.} Each word in question and answer is represented by its $300$-dimension GloVe \cite{DBLP:conf/emnlp/PenningtonSM14} vector.  The sequence of embedded words in question (average length is $9.5$) is then fed into Bi-GRU for each time step.
We have also tried to embed answer with GRU but find that it mostly leads to overfitting since the training set is not huge enough and average answer length is merely $1.2$. So we simply represent the answer by averaging its word embedding.

During training, we utilize Adam optimizer with the mini-batch size as $128$. Dropout and batch normalization are adopted to stabilize the training.
We use a gradual learning rate warm up ($2.5 \times (epoch+1) \times 5 \times 10^{-4}$ ) for the first 7 epochs, decay it at the rate of 0.7 for every 3 epochs for epochs 14 to 47, and remain the same in rest epochs. Meanwhile, the  loss temperature $\tau$ is set to $0.01$ and early stopping is used where $patience$ is equal to  $30$. 
The model is trained offline, and thus the training time usually does not impact the method’s application. In prediction, we currently consider 500 candidate answers for each testing sample. This makes the computation for evaluation affordable. 
\subsection{Overall Results }\label{sec:Overall Results}
\subsubsection{Results on F-VQA.}
\setlength{\tabcolsep}{7pt}
\begin{table}[htbp]
\scriptsize
\caption{The overall results (\% for $Hit@K$) on standard F-VQA datasets (TOP-$500$). $^\dag$ denotes that the model is modified under a mapping-based setting (i.e., remove the last classifier layer of the $(i,q)$ fusion network), which contrasts with traditional classifier-based approach.}
\label{tab:Overall Results of General}
\centering
\begin{tabular}{l|ccccc}
 \toprule
\multirow{1}{*}{{\bf \quad   Methods }}                        
& {\bf $\bm{Hit@1}$}
& $\bm{Hit@3}$
& $\bm{Hit@10}$
& $\bm{MRR}$
& $\bm{MR}$
\\ 
 \midrule
\quad Hie-Q+I \cite{DBLP:conf/nips/LuYBP16}		& $33.70$ & $50.00$  & $64.08$ & - & - \\
\quad MLP			& $34.12$ & $52.26$  & $69.11$ & - & - \\ 	
\quad Up-Down \cite{DBLP:conf/cvpr/00010BT0GZ18} 	& $34.81$ & $50.13$  & $64.37$ & - & - \\
\quad Up-Down$^\dag$ & $40.91$ & $57.47$  & $72.74$  & - & - \\  	
\quad SAN \cite{DBLP:conf/cvpr/YangHGDS16}		& $41.62$ & $58.17$  & $72.69$ & - & - \\
\quad Hie-Q+I+Pre \cite{DBLP:conf/nips/LuYBP16}	& $43.14$ & $59.44$  & $72.20$ & - & - \\	 
\quad BAN \cite{DBLP:conf/nips/KimJZ18}			& $44.02$ & $58.92$  & $71.34 $ & - & - \\ 	 
\quad BAN$^\dag$ & $45.95$ & $63.36$  & $78.12$  & - & - \\
\quad MLP$^\dag$ & $47.55$ & $66.76$  & $81.55$  & - & - \\
\quad SAN$^\dag$ & $49.27$ & $\underline{67.30}$  & $\underline{81.79}$ & 0.605 & 14.75 \\   	 	 	 
\quad top-1-Qqmaping \cite{DBLP:journals/pami/WangWSDH18}	& $52.56$ & $59.72$  & $60.58$ & - & - \\
\quad top-3-Qqmaping \cite{DBLP:journals/pami/WangWSDH18}	& $\underline{56.91}$ & $64.65$  & $65.54$ & - & - \\	 	 
\midrule
\multicolumn{6}{c}{\bf Our Method $(soft~mask~score=10)$}\\
\midrule
\quad $k_r=3,~k_e=1$ & $\bm{58.27}$  & $75.2$   & $86.4$  & $0.683 $ & $11.72$ \\			 	
\quad $k_r=3,~k_e=3$ & $57.42$  & $\bm{76.51}$   & $87.53$  & $\bm{0.685}$ & $10.51$ \\			 	
\quad $k_r=3,~k_e=5$ & $53.84$  & $74.88$   & $\bm{88.49}$  & $0.661$ & $9.58$ \\ 			 	
\quad $k_r=5,~k_e=10$ & $54.02$  & $74.53$   & $88.03$  & $0.660$ & $\bm{9.17}$ \\ 			 	
\bottomrule 
\end{tabular}
\end{table}
To demonstrate the effectiveness of our model under generalized VQA condition, we conduct experiments under standard F-VQA dataset.
Results in Table~\ref{tab:Overall Results of General} gives an overview of the comprehensive evaluation for some representative approaches over this datasets. 
\cjy{It is} interesting that the Up-Down and BAN behave worse than SAN, which may be caused by overfitting of the model due to more parameters and limited training data (less than $3000$). 
But among all those settings, the results demonstrate that our models all outperform corresponding classifier-based or mapping-based models to varying degrees.
The stable improvement (compare with SAN$^\dag$) achieved by our model indicates that adding our method to other end-to-end framework under generalized knowledge VQA  setting could also lead to stable performance improvement. Most importantly, our proposed KG-based framework is independent of fusion model, which makes it possible for multi-scene migration and multi-model replacement.
\subsubsection{Results on ZS-F-VQA.}
We report the prediction results under the standard ZSL setting and GZSL setting in Table~\ref{tab:Overall Results of ZSL}. 
Considering that the traditional classifier-based VQA model fail to work on ZS-VQA since there is no overlap of answer label between the testing set and training set (see Table~\ref{tab:dateset split} for detail), we simply skip these methods here.
We set larger parameters $k$ under ZSL/GZSL setting to mitigate the strong constraint on answer candidate caused by hard mask.
From the overall view, the performance of our model has been significantly improved on the basis of SAN$^\dag$ model. 

Most importantly, the models obtain the state-of-the-art performance under respective indicators with various parameter settings. Take the result in GZSL setting as an example, our method achieve $29.39\%$ improvement  for $hit@1$ (from $0.22\%$ to $29.39\%$), $44.17\%$ for $hit@3$ and $75.34\%$ for $hit@3$.
We have similar observations when the setting transforms into standard ZSL.
To sum up. these observations demonstrate the effectiveness of the model in the ZSL/GZSL scenario, but it also reflects model's dependence on trade off between $k_r$ and $k_e$ (this will be discussed in Section~\ref{sec:Impact of fr}).

\setlength{\tabcolsep}{1pt}
\begin{table}[htbp]
\scriptsize
\caption{The overall results (\% for $Hit@K$) on ZS-F-VQA datasets under the setting of ZSL/GZSL. }
\label{tab:Overall Results of ZSL}
\centering
\begin{tabular}{l|ccccc|ccccc}
\toprule
\multirow{1}{*} & \multicolumn{5}{c}{\bf GZSL} \vline & \multicolumn{5}{c}{\bf ZSL} \\
\midrule
\multirow{1}{*}{{\bf \quad   Methods }}                        
& $\bm{Hit@1}$
& $\bm{Hit@3}$
& $\bm{Hit@10}$
& $\bm{MRR}$
& $\bm{MR}$
& $\bm{Hit@1}$
& $\bm{Hit@3}$
& $\bm{Hit@10}$
& $\bm{MRR}$
& $\bm{MR}$
\\ 
\midrule
\quad Up-Down$^\dag$ & $0.00$ & $2.67$  & $16.48$  & $-$ & $-$ & $13.88$ & $25.87$ & $45.15$ & $-$ & $-$  \\   	 	 	 	 			 
\quad BAN$^\dag$ & $0.22$ & $4.18$  & $18.55$  & $-$ & $-$ & $13.14 $ & $26.92$ & $46.90$ & $-$& $-$ \\  	 	 			 
\quad MLP$^\dag$ & $0.07$ & $4.07$  & $27.40$  & $-$ & $-$ & $18.84$ & $37.85$ & $59.88$ & $-$ & $-$  \\  	 	 	 	 	 
\quad SAN$^\dag$ & $0.11$ & $6.27$  & $31.66$ & $0.093$ & $48.18$ & $20.42$ & $37.20$ & $62.24 $ & $0.337$ & $19.14$  \\	 	 	  	 
\midrule
\multicolumn{11}{c}{\bf ~~~~~~Our Method $(hard~mask~score=100)$}\\
\midrule
\quad $k_r=25,~k_e=1$ & $\bm{29.39}$  & $43.71$   & $62.17$  & $\bm{0.401}$ & $29.52$ & $46.87$ & $62$ & $78.14$ & $0.572$ & $12.22$ \\	 
\quad $k_r=15,~k_e=3$ & $12.22$  & $\bm{50.44}$   & $73.10$  & $0.339 $ & $22.2$ & $\bm{50.51}$ & $70.44$ & $84.24$ & $\bm{0.625}$ & $9.27$  \\			 			 					 	
\quad $k_r=15,~k_e=5$ & $6.69$  & $42.91$   & $\bm{75.34}$  & $0.293$ & $20.61$ & $49.11$ & $\bm{71.17}$ & $86.06$ & $0.622$ & $8.6$  \\				 						 
\quad $k_r=25,~k_e=15$ & $1.96$  & $24.8$   & $72.85$  & $0.208$ & $18.72$ & $40.21$ & $67.04$ & $\bm{88.51}$ & $0.563 $ & $7.68	$ \\  		
\quad $k_r=25,~k_e=25$ & $1.19$  & $18.81$   & $66.97$  & $0.180$ & $\bm{18.14}$ & $35.87$ & $61.86$ & $88.09$ & $0.528$ & $\bm{7.3}$ \\ 	 	 					 	
\bottomrule 
\end{tabular}
\end{table}
\subsection{Ablation Studies} \label{sec:ablation study}
\subsubsection{Choice of Answer Embedding.}
\setlength{\tabcolsep}{8pt}
\begin{table}[htbp]
\scriptsize
\caption{The impact of different answer embedding toward model performance (\%) on standard F-VQA datasets (TOP-$500$). $x(a)$, $h(a)$ and $v(a)$ respectively denote KGE,  ConceptNet embedding, and original GloVe embedding. CLS is classifier-based method.}
\label{tab:Answer Embedding Ablation}
\centering
\begin{tabular}{l|ccc}
 \toprule
\multirow{1}{*}{{\bf \quad   Methods }}                        
& {\bf $\bm{Hit@1}$}
& $\bm{Hit@3}$
& $\bm{Hit@10}$
\\ 
 \midrule
\quad CLS & $38.64$  & $54.87$   & $69.38$   \\			 	
\quad $v(a)$ & $\bm{46.32}$  & $\bm{63.96}$   & $\bm{78.44}$   \\			 	
\quad $x(a)$ & $44.13$  & $59.94$   & $71.94$   \\ 			 	
\quad $h(a)$ & $45.62$  & $62.99$   & $77.34$   \\ 	
\quad $v(a) + h(a)$ & $45.86$  & $63.67$   & $78.43$   \\ 			 	
\quad $v(a) + h(a) + x(a)$ & $45.18$  & $62.95$   & $77.14$   \\ 			 	
\bottomrule 
\end{tabular}
\end{table}
To compare the influence of answer embedding in feature projection performance, we define $g_{\phi}(a)=g_{\phi}(\mathcal{C}[x(a);h(a);v(a)])$ in this part where $\mathcal{C}$ denotes simple concatenate function. Specially, $x(a)$, $h(a)$ and $v(a)$ respectively denotes KG embedding (KGE),  ConceptNet embedding \cite{DBLP:conf/aaai/MalaviyaBBC20}, and original GloVe embedding. 
This KGE technique can be used to complete the KG with missing entities or links, meanwhile produce the embedding of nodes and links as their representations. 
Specially, we adopt TransE \cite{DBLP:conf/nips/BordesUGWY13} as $x(.)$ and train it on our KG. 
As for $h(a)$, we utilize the BERT-based node representations generated by a pre-trained common sense embedding model \cite{DBLP:conf/aaai/MalaviyaBBC20}, which exploits the structural and semantic context of nodes within a large scale common sense KG.
As the result shown in Table~\ref{tab:Answer Embedding Ablation}, when work independently, word2vec representation ($78.44\%$) of answers exceed graph based methods ( $71.94\%$ for KGE and $77.34\%$ for ConceptNet Embedding in $Hit@10$ ) in performance even though they contain more information. 
We guess that when the size of the dataset is small, the complexity of neural network limits model's sensitivity to the input representation. So finally we simply choose GloVe as the initial representation of all inputs.
 \subsubsection{Impact of Mask Score.} \label{sec:Impact of mask score}
\begin{figure}[htbp]
\centering
\includegraphics[width=0.3\textwidth]{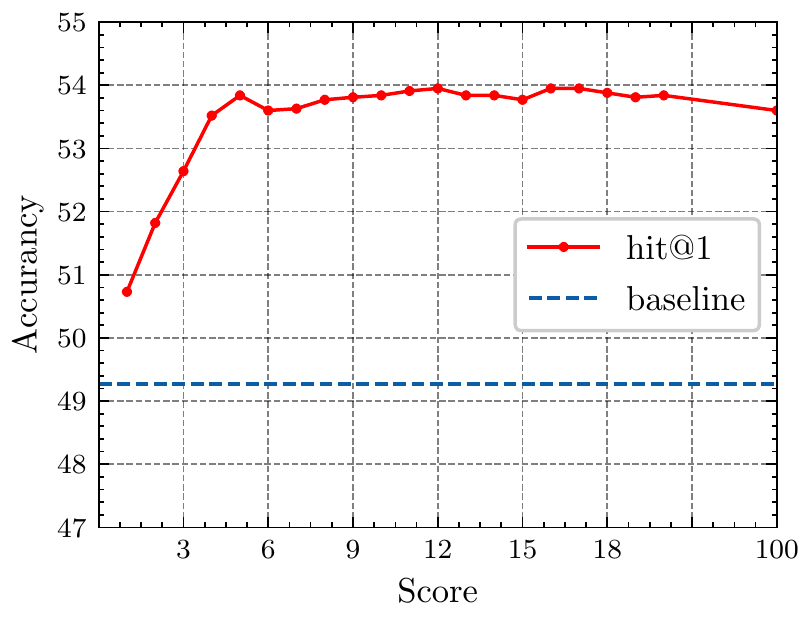} 
\includegraphics[width=0.3\textwidth]{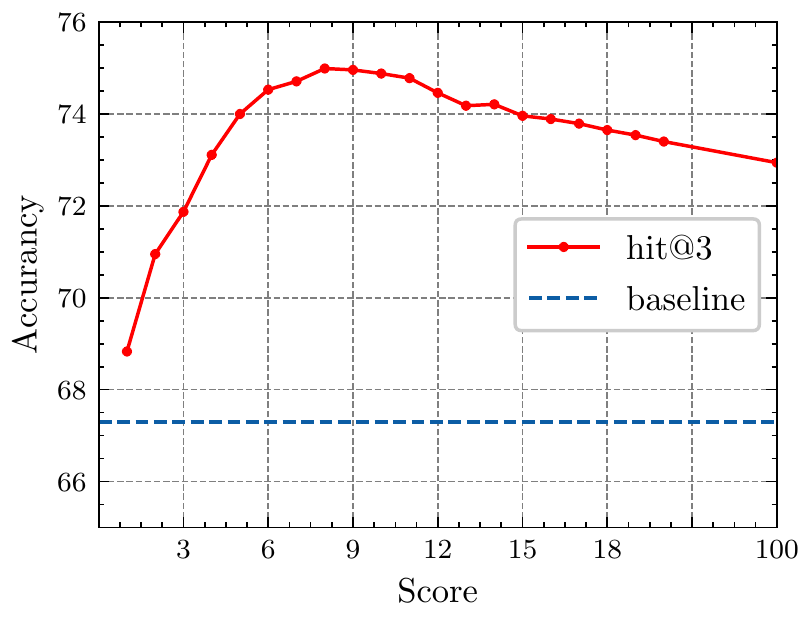}
\includegraphics[width=0.3\textwidth]{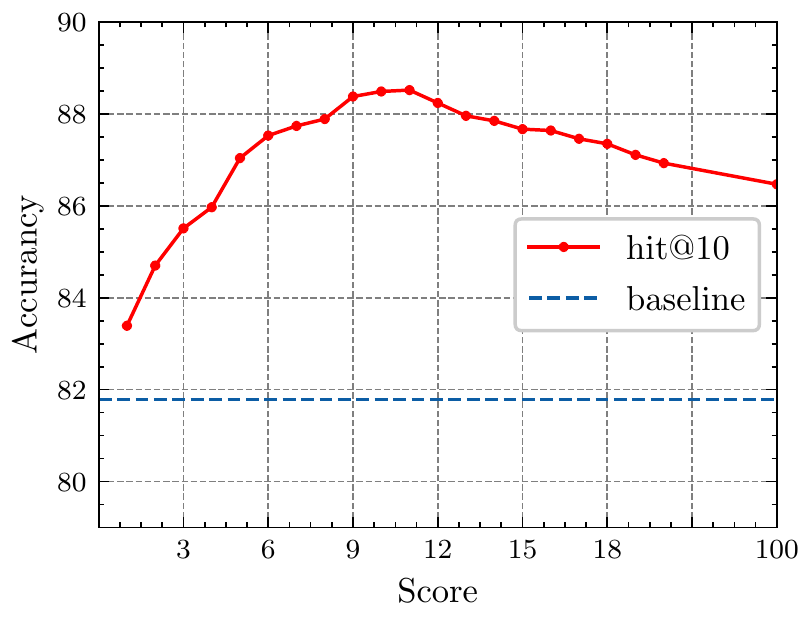}

\includegraphics[width=0.3\textwidth]{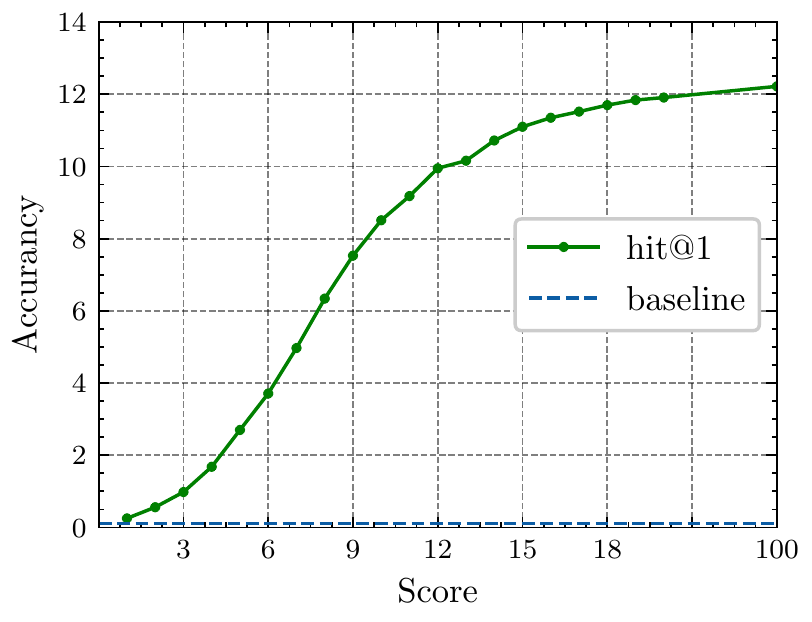} 
\includegraphics[width=0.3\textwidth]{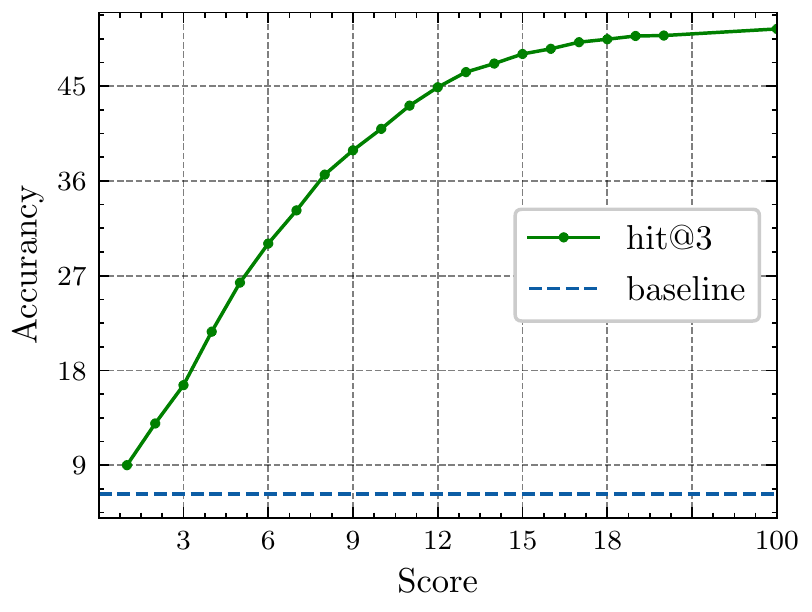}
\includegraphics[width=0.3\textwidth]{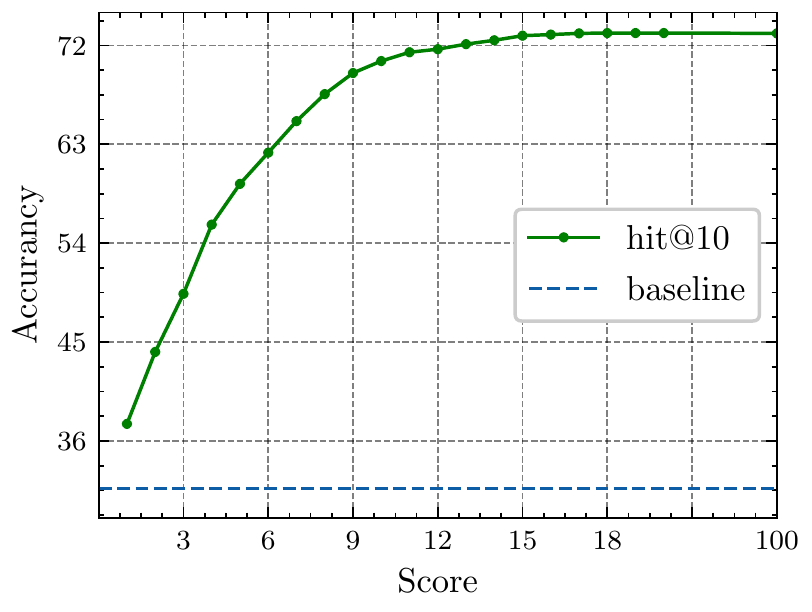}
\caption{Impact of mask score in standard F-VQA ($k_r = 3$, $k_e = 10$) under generalized setting (Up), and ZS-F-VQA ($k_r = 15$, $k_e = 3$) under GZSL setting (Down).}
\label{fig:score in F-VQA}
\label{fig:score in ZS-F-VQA}
\end{figure}

In this part we mainly discuss the effect of mask score on ZS-F-VQA and F-VQA which is reflected by $hit@1$ (Left), $hit@3$ (Middle) and $hit@10$ (Right) accuracy as shown in Figure~\ref{fig:score in ZS-F-VQA}.
Caused by the sparsity of high-dimensional space vector, the value of $F_{\boldsymbol{\theta\diamond}}(i, q)^{\top} G_{\boldsymbol{\phi\diamond}}(f_n))$ is quite small as we observing on experiment. 
This is also another reasons why we define $\tau$ for the scale-up of vector similarity (in addition to accelerating model convergence).
Considering that $sim(\left(i,q),a_n\right)$ distributes from $145$ to $232$, we simply take $100$ as the dividing line of score between hard mask and soft mask which is big enough for correcting an incorrect answer into a correct one in testing stage.
As shown in Figure~\ref{fig:score in F-VQA}, the result gaps between soft mask (i.e., low score) and hard mask (i.e., high score)  are completely different in standard and GZSL VQA scenarios.
We consider following reasons:
1) Firstly, do not try to rely on network to model complex common sense knowledge when data is scarce: 
When applied to ZS-F-VQA, we notice that model merely learns prior shallow knowledge representation and poor transfer capabilities for unseen answers (see Section~\ref{sec:Interpretability}). In this case, the strong guiding capability of additional knowledge makes a great contribution to answer prediction.
2) Secondly, if the training samples are sufficient, the error cascading caused by pipeline mode may become the restriction of model performance: 
When applied to standard F-VQA, the model itself already has high confidence in correct answer and external knowledge should appropriately relax the constraint. 
We observe that overly strong guidance (i.e., hard mask) becomes a burden at this moment, so soft mask is in demand as a soft constraint.
This reflects the necessity of defining different mask.
\subsubsection{Impact of Support Entity and Relation.}\label{sec:Impact of fr}
\begin{figure}[htbp]
\centering
\includegraphics[width=0.32\textwidth]{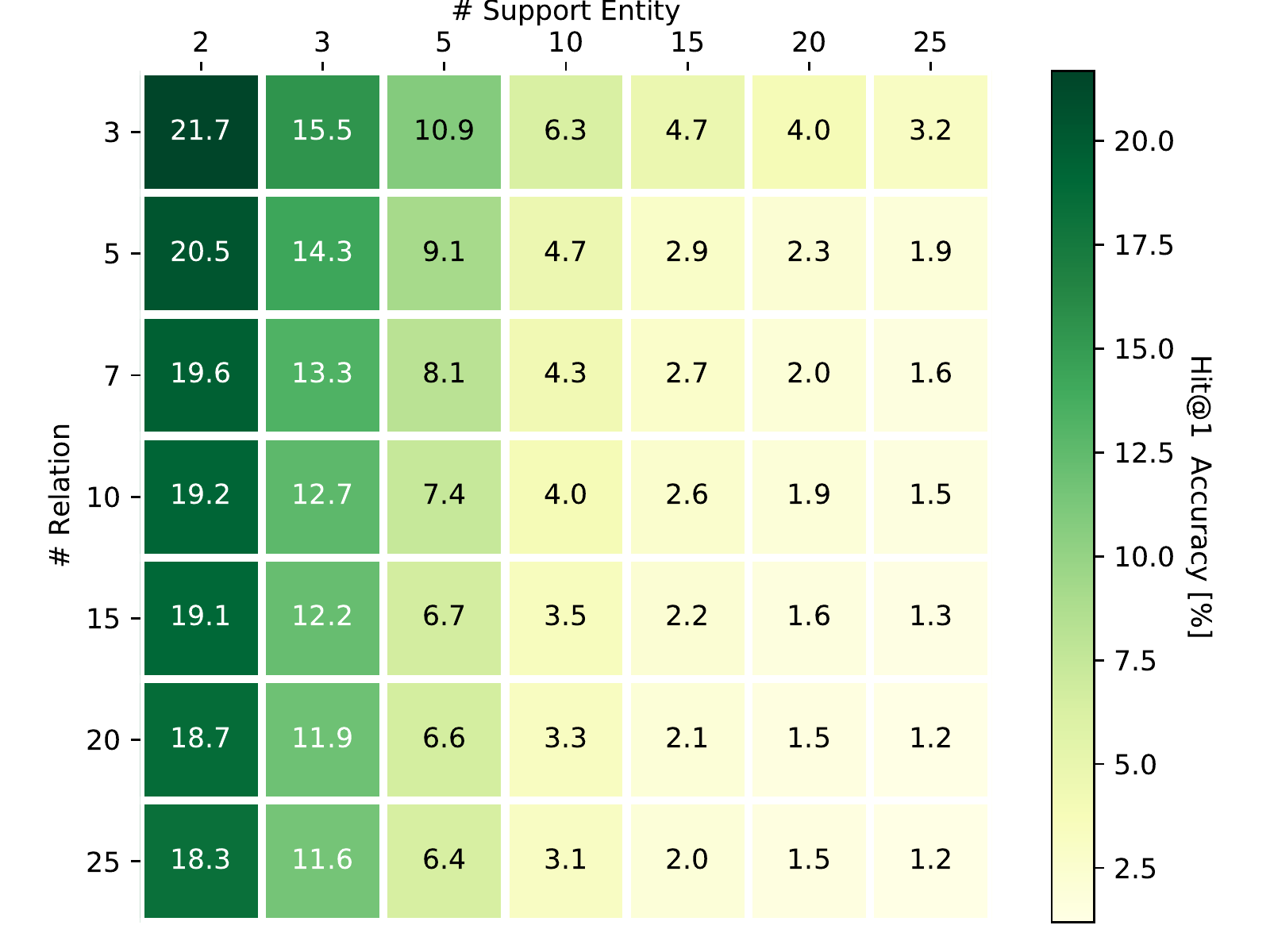} 
\includegraphics[width=0.32\textwidth]{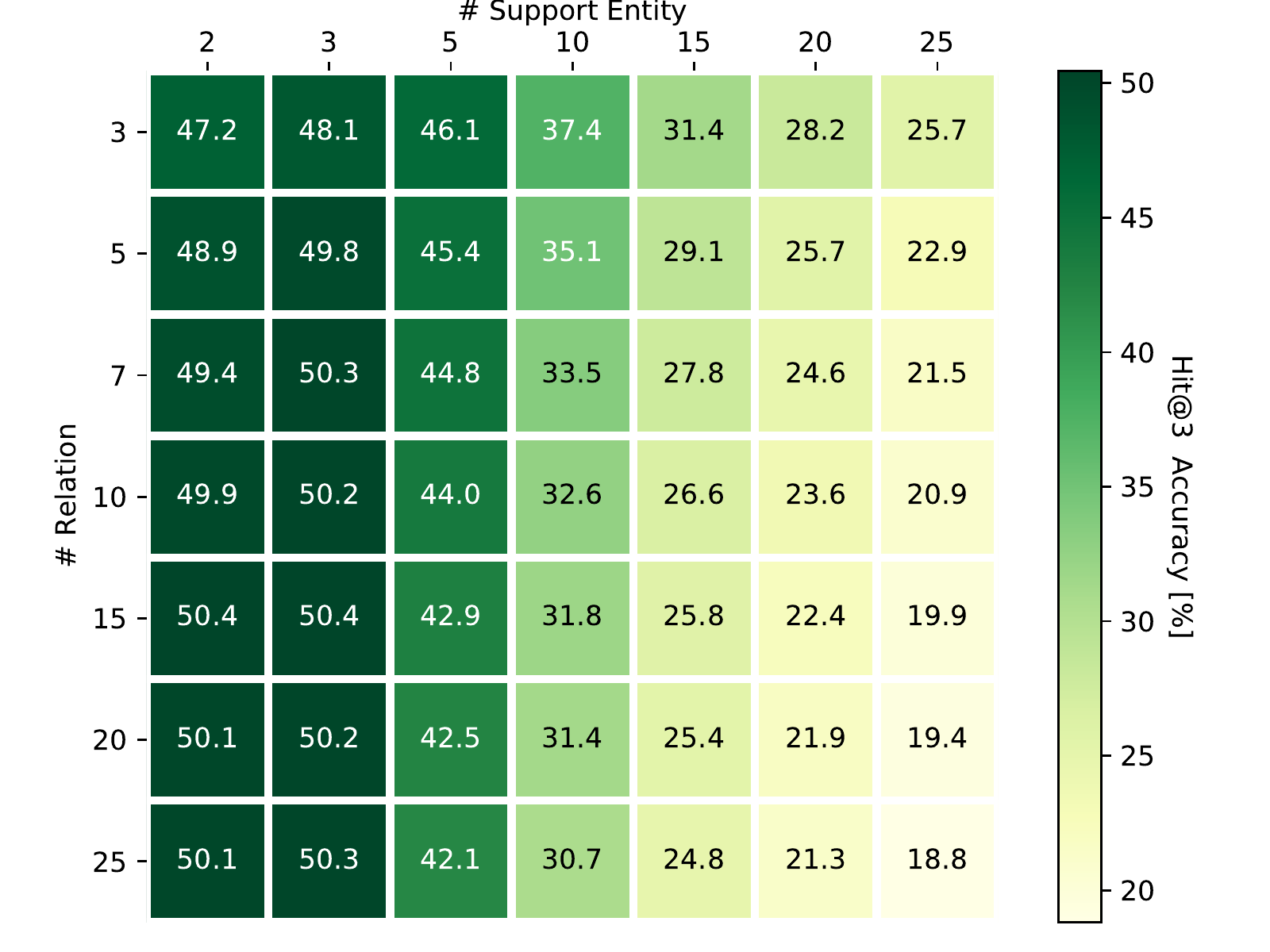}
\includegraphics[width=0.32\textwidth]{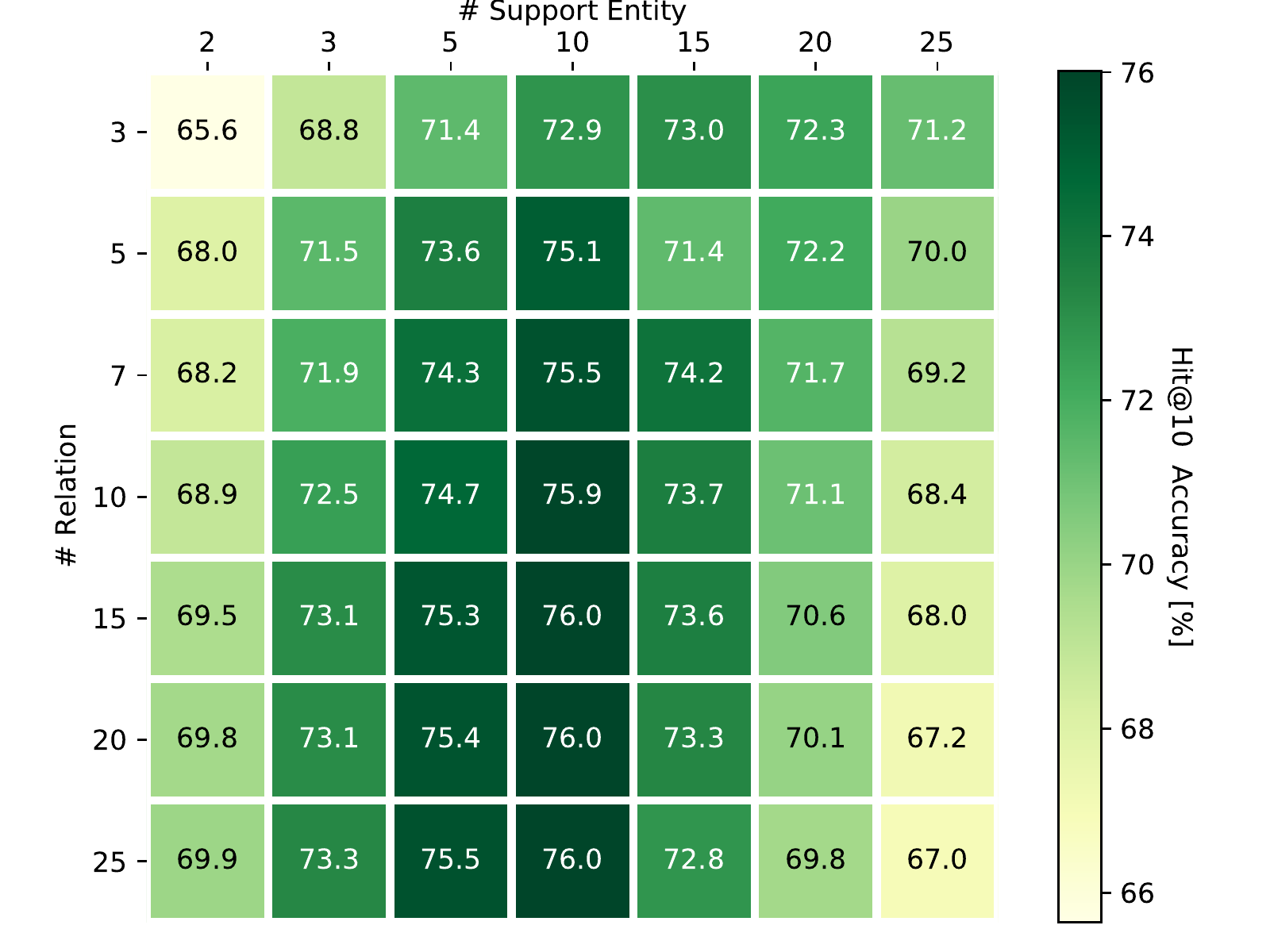}
\caption{Impact of \#support entity ($k_e$) and \#relation ($k_r$) on GZSL setting.}
\label{fig:heatmap of k}
\end{figure}

As shown in Figure~\ref{fig:heatmap of k}, we notice that $hit@1$ and $hit@10$ cannot simultaneously achieve the best, despite that the model can always  exceed the baseline a lot with different $k$. This phenomenon is plausible since that the more restrictive target candidate set is, the more likely it succeed predicting answer in a smaller range, with the cost of missing some other true answers due to the error prediction of support entity/relation. The contrast between MRR and MR well reflects this view (see Table~\ref{tab:Overall Results of ZSL}).
\subsection{Interpretability}\label{sec:Interpretability}
\begin{figure}[htbp]
\centering
\includegraphics[width=1.0\textwidth]{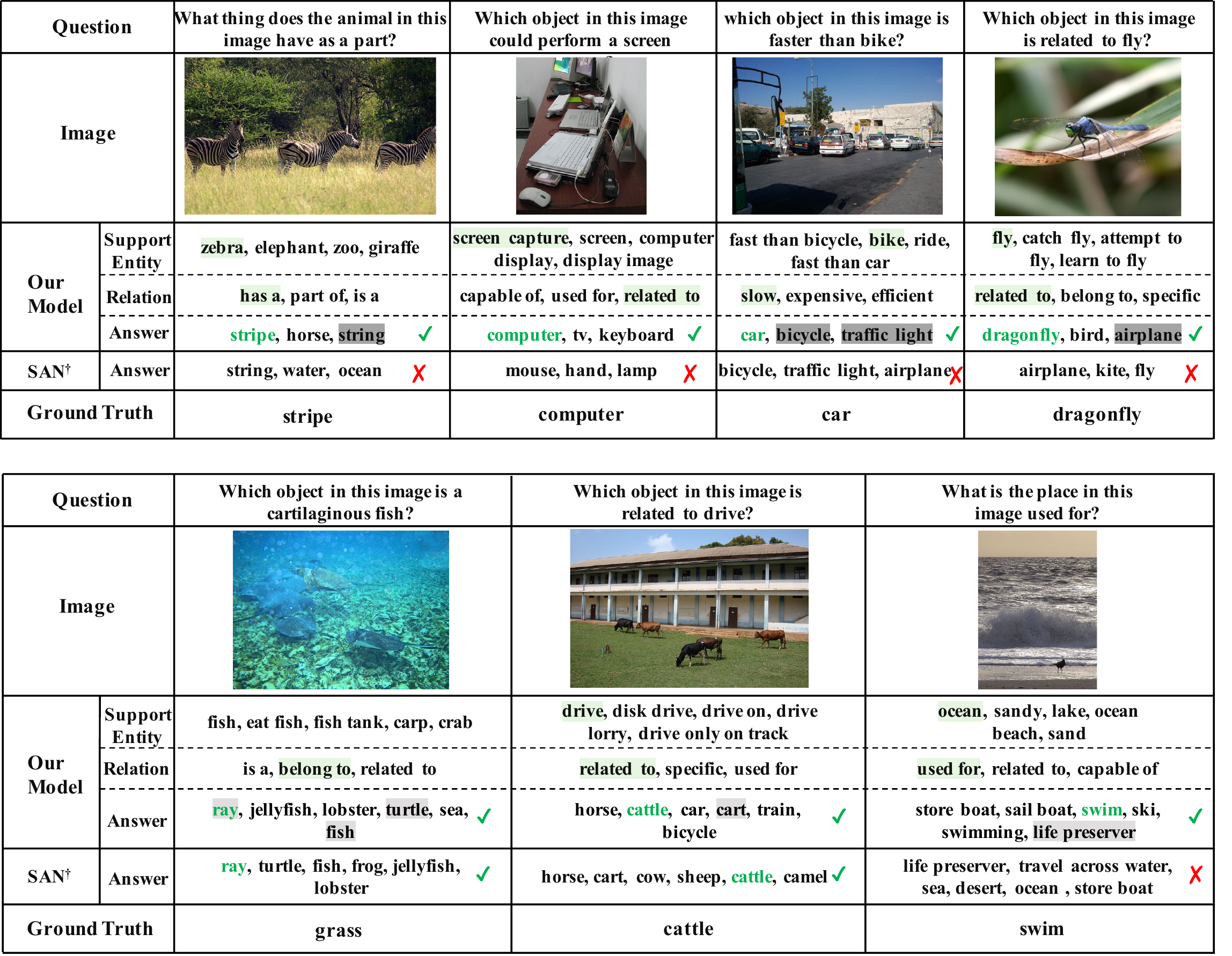} 
\caption{Cases under GZSL VQA (Up) and Generalized VQA (Down) setting. }
\label{fig:example of zsl}
\label{fig:example of general}
\end{figure}

To further validate the effectiveness of our knowledge-based ZS-VQA model, we visualize the output and intermediate process of our method compared to best baseline model SAN$^\dag$ \cite{DBLP:conf/cvpr/HuCS18}.
Figure~\ref{fig:example of zsl} (Up) shows the detected support entities, relations, and answers for four examples in ZS-F-VQA dataset together with answer predicted by SAN$^\dag$ and the groudtruth one. 
It indicates that normal models tend to align answer directly with meaning content in question/image (e.g. bicycle in Case 3) or match the seen answers (e.g. airplane in case 4 ), which is a lazy way of learning accompanied by overfitting.  
To some extend, the difficult common sense knowledge stored in structured data is utilized to playing a guiding role here.
Our method can also be generalized to predict multiple answers since the probabilistic model can calculate scores for all candidates to select the top-K answers (see answer ``tv" in Case 2 of Figure \ref{fig:example of zsl}).

Our method also works well under generalized VQA setting as illustrated in Figure~\ref{fig:example of general} (Down). For those simpler answers, it can increase the probability (e.g. Case $6$) for correct prediction. More importantly, distinguish from the hard mask (dark shadows) in ZSL setting, the soft mask strategy here effectively alleviates error cascading which reduces the influence from previous model's error (e.g. failed prediction on support entity lead to the error mask on Case $5$).
\section{Conclusion}
We propose a Zero-shot VQA model via knowledge graph for addressing the problem of exploiting external knowledge for Zero-shot VQA. 
\cjy{The crucial factor} to the success of our method is to consider both the knowledge contained in \cjy{the} answer itself and the external common sense knowledge from knowledge graphs. Meanwhile we convert VQA from a traditional classification task to a mapping-based alignment task for \cjy{addressing} unseen answer prediction.
Experiments on multiple models support our claim that this method \cjy{can not only} achieve outstanding performance in zero-shot scenarios but also make steady progress at different end-to-end models on \cjy{the general VQA task}. 
Next we will further investigate KG construction and KG embedding \cjy{methods for more robust but compact semantics for addressing ZS-VQA. }
Moreover, we will release and improve the ZS-VQA codes and data, in conjunction with K-ZSL \cite{DBLP:journals/corr/abs-2106-15047}.

\subsubsection{Acknowledgments.}
This work is funded by 2018YFB1402800/NSFCU19B2027 /NSFC91846204.
Jiaoyan Chen is founded by the SIRIUS Centre for Scalable Data Access (Research Council of Norway) and Samsung Research UK.


%
%

\bibliographystyle{splncs04}
\bibliography{references}
\begin{thebibliography}{10}
\providecommand{\url}[1]{\texttt{#1}}
\providecommand{\urlprefix}{URL }
\providecommand{\doi}[1]{https://doi.org/#1}

\bibitem{DBLP:journals/corr/AgrawalKBP17}
Agrawal, A., Kembhavi, A., Batra, D., Parikh, D.: {C-VQA:} {A} compositional
  split of the visual question answering {(VQA)} v1.0 dataset. CoRR
  \textbf{abs/1704.08243} (2017)

\bibitem{DBLP:conf/cvpr/00010BT0GZ18}
Anderson, P., He, X., Buehler, C., Teney, D., Johnson, M., Gould, S., Zhang,
  L.: Bottom-up and top-down attention for image captioning and visual question
  answering. In: {CVPR}. pp. 6077--6086 (2018)

\bibitem{DBLP:conf/iccv/AntolALMBZP15}
Antol, S., Agrawal, A., Lu, J., Mitchell, M., Batra, D., Zitnick, C.L., Parikh,
  D.: {VQA:} visual question answering. In: {ICCV}. pp. 2425--2433 (2015)

\bibitem{DBLP:conf/nips/BordesUGWY13}
Bordes, A., Usunier, N., Garc{\'{\i}}a{-}Dur{\'{a}}n, A., Weston, J.,
  Yakhnenko, O.: Translating embeddings for modeling multi-relational data. In:
  {NIPS}. pp. 2787--2795 (2013)

\bibitem{chen2021knowledge}
Chen, J., Geng, Y., Chen, Z., Horrocks, I., Pan, J.Z., Chen, H.:
  Knowledge-aware zero-shot learning: Survey and perspective. In: IJCAI Survey
  Track (2021)

\bibitem{DBLP:conf/icml/ChenG0LC020}
Chen, L., Gan, Z., Cheng, Y., Li, L., Carin, L., Liu, J.: Graph optimal
  transport for cross-domain alignment. In: {ICML}. vol.~119, pp. 1542--1553
  (2020)

\bibitem{DBLP:conf/emnlp/ChenZCXWW20}
Chen, W., Zha, H., Chen, Z., Xiong, W., Wang, H., Wang, W.Y.: Hybridqa: {A}
  dataset of multi-hop question answering over tabular and textual data. In:
  {EMNLP}. pp. 1026--1036 (2020)

\bibitem{DBLP:conf/naacl/DevlinCLT19}
Devlin, J., Chang, M., Lee, K., Toutanova, K.: {BERT:} pre-training of deep
  bidirectional transformers for language understanding. In: {NAACL}. pp.
  4171--4186 (2019)

\bibitem{DBLP:journals/ivc/FaraziKB20}
Farazi, M.R., Khan, S.H., Barnes, N.: From known to the unknown: Transferring
  knowledge to answer questions about novel visual and semantic concepts. Image
  Vis. Comput.  \textbf{103},  103985 (2020)

\bibitem{DBLP:conf/www/GengC0PYYJC21}
Geng, Y., Chen, J., Chen, Z., Pan, J.Z., Ye, Z., Yuan, Z., Jia, Y., Chen, H.:
  Ontozsl: Ontology-enhanced zero-shot learning. In: {WWW}. pp. 3325--3336
  (2021)

\bibitem{DBLP:journals/corr/abs-2106-15047}
Geng, Y., Chen, J., Chen, Z., Pan, J.Z., Yuan, Z., Chen, H.: {K-ZSL:} resources
  for knowledge-driven zero-shot learning. CoRR  \textbf{abs/2106.15047} (2021)

\bibitem{DBLP:conf/cvpr/HuCS18}
Hu, H., Chao, W., Sha, F.: Learning answer embeddings for visual question
  answering. In: {CVPR}. pp. 5428--5436 (2018)

\bibitem{DBLP:conf/nips/KimJZ18}
Kim, J., Jun, J., Zhang, B.: Bilinear attention networks. In: NeurIPS. pp.
  1571--1581 (2018)

\bibitem{DBLP:conf/nips/LuYBP16}
Lu, J., Yang, J., Batra, D., Parikh, D.: Hierarchical question-image
  co-attention for visual question answering. In: {NIPS}. pp. 289--297 (2016)

\bibitem{DBLP:conf/aaai/MalaviyaBBC20}
Malaviya, C., Bhagavatula, C., Bosselut, A., Choi, Y.: Commonsense knowledge
  base completion with structural and semantic context. In: {AAAI}. pp.
  2925--2933 (2020)

\bibitem{DBLP:conf/cvpr/MarinoRFM19}
Marino, K., Rastegari, M., Farhadi, A., Mottaghi, R.: {OK-VQA:} {A} visual
  question answering benchmark requiring external knowledge. In: {CVPR}. pp.
  3195--3204 (2019)

\bibitem{DBLP:conf/nips/NarasimhanLS18}
Narasimhan, M., Lazebnik, S., Schwing, A.G.: Out of the box: Reasoning with
  graph convolution nets for factual visual question answering. In: NeurIPS.
  pp. 2659--2670 (2018)

\bibitem{DBLP:conf/eccv/NarasimhanS18}
Narasimhan, M., Schwing, A.G.: Straight to the facts: Learning knowledge base
  retrieval for factual visual question answering. In: {ECCV} {(8)}. vol.
  11212, pp. 460--477 (2018)

\bibitem{PCEH+2017}
Pan, J., Calvanese, D., Eiter, T., Horrocks, I., Kifer, M., Lin, F., Zhao, Y.
  (eds.): {Reasoning Web: Logical Foundation of Knowledge Graph Construction
  and Querying Answering}. Springer (2017)

\bibitem{PVGW2017}
Pan, J., Vetere, G., Gomez-Perez, J., Wu, H. (eds.): {Exploiting Linked Data
  and Knowledge Graphs for Large Organisations}. Springer (2017)

\bibitem{DBLP:conf/emnlp/PenningtonSM14}
Pennington, J., Socher, R., Manning, C.D.: Glove: Global vectors for word
  representation. In: {EMNLP}. pp. 1532--1543 (2014)

\bibitem{DBLP:conf/cvpr/RamakrishnanPSM17}
Ramakrishnan, S.K., Pal, A., Sharma, G., Mittal, A.: An empirical evaluation of
  visual question answering for novel objects. In: {CVPR}. pp. 7312--7321
  (2017)

\bibitem{DBLP:conf/aaai/ShahMYT19}
Shah, S., Mishra, A., Yadati, N., Talukdar, P.P.: {KVQA:} knowledge-aware
  visual question answering. In: {AAAI}. pp. 8876--8884 (2019)

\bibitem{DBLP:journals/corr/abs-2005-01239}
Shevchenko, V., Teney, D., Dick, A.R., van~den Hengel, A.: Visual question
  answering with prior class semantics. CoRR  \textbf{abs/2005.01239} (2020)

\bibitem{DBLP:journals/corr/TeneyH16a}
Teney, D., van~den Hengel, A.: Zero-shot visual question answering. CoRR
  \textbf{abs/1611.05546} (2016)

\bibitem{DBLP:conf/ijcai/WangWSDH17}
Wang, P., Wu, Q., Shen, C., Dick, A.R., van~den Hengel, A.: Explicit
  knowledge-based reasoning for visual question answering. In: {IJCAI}. pp.
  1290--1296 (2017)

\bibitem{DBLP:journals/pami/WangWSDH18}
Wang, P., Wu, Q., Shen, C., Dick, A.R., van~den Hengel, A.: {FVQA:} fact-based
  visual question answering. {IEEE} TPAMI  \textbf{40}(10),  2413--2427 (2018)

\bibitem{DBLP:conf/cvpr/WuWSDH16}
Wu, Q., Wang, P., Shen, C., Dick, A.R., van~den Hengel, A.: Ask me anything:
  Free-form visual question answering based on knowledge from external sources.
  In: {CVPR}. pp. 4622--4630 (2016)

\bibitem{DBLP:conf/cvpr/YangHGDS16}
Yang, Z., He, X., Gao, J., Deng, L., Smola, A.J.: Stacked attention networks
  for image question answering. In: {CVPR}. pp. 21--29 (2016)

\bibitem{DBLP:journals/corr/abs-1908-06725}
Ye, Z., Chen, Q., Wang, W., Ling, Z.: Align, mask and select: {A} simple method
  for incorporating commonsense knowledge into language representation models.
  CoRR  \textbf{abs/1908.06725} (2019)

\bibitem{DBLP:conf/ijcai/ZhuYWS0W20}
Zhu, Z., Yu, J., Wang, Y., Sun, Y., Hu, Y., Wu, Q.: Mucko: Multi-layer
  cross-modal knowledge reasoning for fact-based visual question answering. In:
  {IJCAI}. pp. 1097--1103 (2020)

\end{thebibliography}
\end{document}